\pdfoutput=1
\documentclass[11pt]{article}

\usepackage[final]{acl}

\usepackage{times}
\usepackage{latexsym}
\usepackage{graphicx, tabularx}
\usepackage{booktabs}
\usepackage{multirow}
\usepackage{enumitem}
\usepackage{amsmath}

\usepackage{url}

\usepackage{breakurl}
\usepackage[breaklinks]{hyperref}

\usepackage[T1]{fontenc}
\usepackage[utf8]{inputenc}

\usepackage{microtype}

\usepackage{inconsolata}

\DeclareMathAlphabet\mathbfcal{OMS}{cmsy}{b}{n}

\title{Whose Preferences? Differences in Fairness Preferences and Their Impact on the Fairness of AI Utilizing Human Feedback}
  
\author{Emilia Agis Lerner \\
  ETH Zürich \\
  \texttt{eagislerner@ethz.ch} \\\And
  Florian E. Dorner \\
  MPI for Intelligent Systems, Tübingen \\ ETH Zürich \\
  \texttt{florian.dorner@tuebingen.mpg.de} \\ \AND
  Elliott Ash \\
  ETH Zurich \\
  \texttt{ashe@ethz.ch} \\ \And
  Naman Goel \\
    University of Oxford  \\
\texttt{naman.goel@cs.ox.ac.uk}
  }

\begin{document}
\maketitle
\begin{abstract}
There is a growing body of work on learning from human feedback to align various aspects of machine learning systems with human values and preferences. We consider the setting of fairness in content moderation, in which human feedback is used to determine how two comments --- referencing different sensitive attribute groups --- should be treated in comparison to one another. With a novel dataset collected from Prolific and MTurk, we find significant gaps in fairness preferences depending on the race, age, political stance, educational level, and LGBTQ+ identity of annotators. We also demonstrate that demographics mentioned in text have a strong influence on how users perceive individual fairness in moderation.  Further, we find that differences also exist in downstream classifiers trained to predict human preferences. Finally, we observe that an ensemble, giving equal weight to classifiers trained on annotations from different demographics, performs better for different demographic intersections; compared to a single classifier that gives equal weight to each annotation.

\textcolor{red}{\textit{Warning: This paper discusses examples of content that may be offensive or disturbing.}}
\end{abstract}

\section{Introduction}
With artificial intelligence (AI) and machine learning (ML) systems being deployed in wide-ranging scenarios, there are growing calls for safe, responsible and trustworthy development of these systems~\cite{executive-order,ai-act-eu,bletchley-declaration}. An idea in this space that is receiving a lot of attention lately is that of learning from human feedback for making these systems more aligned with human values and preferences~\cite{ci-anthropic23, bai2022training}. The idea is certainly promising, as human preferences about values like fairness can often not be expressed as simple mathematical constraints~\cite{jung2021algorithmic,florian}, but there also remain several important open questions about the idea~\cite{kirk-etal-2023-past}.

\begin{figure}
\centering
\includegraphics[width=1\columnwidth]{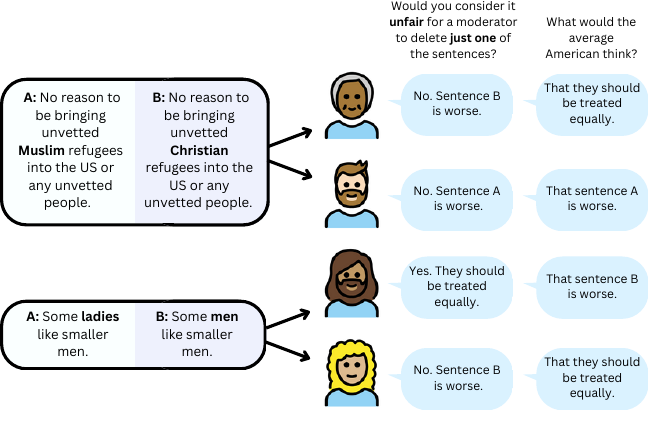}
\caption{Example tasks in our survey, asking people about their fairness preferences and their guess of the average American answer. Each task contains a pair of sentences; sentences in a pair differ along a sensitive attribute such as religion, gender, etc.}
\label{fig:approach}
\end{figure}

In this paper, we focus on the process of eliciting and aggregating human feedback for learning fairness preferences. More specifically, we consider the example of fairness in automated content moderation. Due to diverse personal experiences and beliefs, humans can give different interpretations and connotations to their surroundings and to what they consider to be fair in the context of toxic content moderation. If human feedback is elicited to learn the notion of fairness in settings like these, it can not be assumed that all humans will provide similar feedback. Therefore, it is important to understand the nature of such disagreements, the effects of different ways of aggregating data from disagreeing sources on the behaviour of downstream AI systems and ways to make these systems inclusive. 

While previous work has studied the effect of annotators' demographic identities in toxicity labels (discussed further in Section~\ref{sec:related}), our work focuses on individual fairness preferences, i.e. whether or not two entities ought to receive similar treatment~\citealp{dwork2012fairness}. We collect data from online crowdsourcing platforms (\href{https://www.mturk.com}{MTurk} and \href{https://www.prolific.com}{Prolific}) to understand these issues. Similar to \citealp{florian}, we present participants with pairs of semantically similar -often toxic- comments (in English language), each of which references a different sensitive attribute group. For each pair, we collect participants' opinions on how both sentences should be treated by a content moderator in comparison to one another. In addition, we ask participants to predict the "average American" answer to the same question.\footnote{Please see Section \ref{sec:limitations} for a discussion about the limitations of using the term "average American" in our survey.} Thus, human judgments in our data do not refer to the toxicity level of single sentences, but rather to the fair treatment of two comments relative to each another. \autoref{fig:approach} shows our survey approach with some examples.

Using the collected data, we study how annotators' demographic characteristics (gender, race, age, political stance, educational level, and LGBTQ+ identity) influence their fairness judgments. In addition, we also study the effects of references to different sensitive attribute groups in the sentence pairs. We find significant differences in fairness preferences depending on the demographic identities of annotators, as well as the sensitive attribute mentioned in the sentence pairs. Further, we show that training on different annotator groups' preference data significantly impacts the behaviour of downstream models trained to predict fairness preferences. Perhaps surprisingly, models that are trained on data from a given annotator demographic group $a$ do not usually outperform different demographic groups' models when evaluated on data from $a$. Instead, models perform better when evaluated on data from certain annotator groups like age 38+, no matter which group's data they were trained on.

Finally, motivated by prior work on dealing with disagreements in crowdsourcing (e.g., ~\citealp{davani-etal-2022-dealing,gordon2022jury}), we investigate whether ensembling models for different demographics can improve predictions of fairness preferences for intersections of demographic groups. We find that ensembling does help in this case too, perhaps by reducing the impact of skewness in the number of labels provided by different groups. While fundamental tensions remain, this approach can provide more representation to the marginalised groups. 

\paragraph{Availability of Data} The dataset collected for this work is of independent interest for future research in this area. The crowdsourced data is available at \url{https://github.com/EmiliaAgis/Differences-in-Fairness-Preferences-ACL-2024}. Please note that the dataset contains examples of sentences that may be offensive and disturbing. Personal identifiable information of the annotators has been removed. We also provide a readme file and a license file in the repository.

\section{Related Work}\label{sec:related}
\paragraph{Individual Fairness in Machine Learning} \citealp{dwork2012fairness} describe individual fairness as the principle that two individuals or instances that are similar with regards to a task, should be treated similarly. This is operationalised in terms of Lipschitz continuity of the mapping from instances to treatments, which can be shown to reduce to robustness to a metric-dependent set of perturbations for binary classification \cite{florian}. For example, it would likely seem unfair for a content moderator to remove a comment written in African-American Vernacular English \cite{rios2020fuzze}, while not removing a semantically equivalent comment written in Standard American English. \citealp{garg2019counterfactual} operationalise this idea regarding the referent rather than the writer of a text. They suggest to measure fairness by comparing model outputs for a -often toxic- sentence $s$ and versions $s'$ of the same sentence that have been modified by replacing words on a list of demographic identifiers with each other. \citealp{florian} extend this beyond word replacement by using large language models for unsupervised rewriting in terms of demographic references. Human feedback is then used to validate whether these  generated pairs align with human intuitions on individual fairness. The pairs are then used to train a classifier that predicts fairness judgments, which can be used as fairness constraints for downstream models. In this work, we collect annotators' fairness judgments on the pairs generated by \citealp{florian} and investigate how annotators' demographics affect their fairness judgments. 

\paragraph{Influence of Demographics on Human Annotations} \citealp{goyal2022your,larimore-etal-2021-reconsidering,kumar2021designing} explore the effect of race and LGBTQ+ identity on the perception of negativity in online comments. The first finds that annotators are more sensitive when exposed to toxic comments that regard their own minority group, while the latter two find that minority annotators react more negatively to certain topics like police brutality and are more likely to flag comments as harassment. Other work has found differences of varying strengths between classifiers for toxic content that have been trained on data provided by annotators from different demographics \cite{al2020identifying,binns2017like}. Meanwhile, \citealp{fleisig2023majority} found that providing hate speech classifiers with additional information about annotator demographics greatly improves their ability to predict \textit{individual} annotator judgments. Instead of predicting the majority label, predictions of these individual annotator can be combined to combat majoritarian biases in downstream applications \cite{fleisig2023majority,gordon2022jury}. While our analysis is inspired by these works and others such as~\cite{sap2022annotators}, we are the first to study the effects of demographics on individual fairness preferences rather than the perception of toxicity. 

\paragraph{Variability and Disagreement in Crowdsourced Data} There is also relevant literature~\cite{plank2022problem} on variation and disagreement in data collected from a crowd. This literature includes (but is not limited to)~\cite{aroyo2015truth,inel2014crowdtruth,aroyo2019crowdsourcing,arhin2021ground,denton2021whose,fleisig2023majority,aroyo2024dices}. Unlike prior work, we focus on differences in individual fairness preferences and the effects on downstream classifiers. More recently, there has been work on exploring whether the large language models reflect the "opinions" of certain populations more than that of others~\cite{santurkar2023whose,durmus2023towards,ryan2024unintended}.

\section{Data}\label{sec:data}

\paragraph{Dataset from Prior Work} Comment pairs from \citealp{florian} consist of a large set of often toxic social media comments from the \href{https://www.kaggle.com/competitions/jigsaw-unintended-bias-in-toxicity-classification/data}{Jigsaw Civil Comments dataset} that mention different demographic groups, as well as automatically generated semantically similar versions of these comments that reference a different demographic group than the original comment. For a subset (6902) of these comment pairs, annotations from crowdworkers are also available. We will refer to this annotated dataset as \textbf{$\mathcal{D}_{o}$}. Each of the pairs in $\mathcal{D}_{o}$ has one to nine labels on annotators' predictions on whether or not the average American believes that both comments ought to be treated the same. 720 sentence pairs in $\mathcal{D}_{o}$ also contain nine annotators' \textit{personal} opinions on the fair treatment of the sentence pairs. Further, in $\mathcal{D}_{o}$, annotators could label as many comment pairs as they liked, potentially leading to a few active annotators providing many or most labels.

\paragraph{New Dataset} To remedy the above issues with $\mathcal{D}_{o}$, we construct a new dataset for our research. We first select a subset of comment pairs from $\mathcal{D}_{o}$. As we are interested in investigating disagreements between different annotators, we focus on sentence pairs with particularly contentious fairness judgments: After manually filtering out low quality pairs, for which the original and modified sentences $s$ and $s'$ are not semantically similar, we collect labels for the 1500 sentence pairs with the largest \textit{unlikeability} ~\cite{kader2007variability,perry2005variation}, a measure for the variability in categorical variables. Unlikeability is defined as $u_2(s,s') = 1 - \sum_i p_i^2 \in [0,1-\frac{1}{i}]$ where $p_i$ is the relative frequency of each of the $i$ annotation options for $s,s'$. Clearly $u_2(s,s')=0$ if and only if responses are unanimous, and large unlikeability coefficients correspond to high variability in ratings. Further details about our selection of sentence pairs to label can be found in Appendix \ref{appendix:selection}. 

We then collect 1000 annotators' personal fairness judgments as well as their predictions for the average American's judgment, and data on annotators' gender, race, age, political stance, education, and LGBTQ+ identity using MTurk and Prolific, with 500 annotators per platform. Each annotator labeled 20 sentence pairs, as well as an attention check question, and each sentence pair was labeled by at least 10 annotators. We provide annotators with the following four options to select from:
\begin{enumerate}[topsep=0.5pt,itemsep=0.01ex,partopsep=0ex,parsep=0.5ex]
    \item It would be unfair to delete one of the comments, but not the other.
    \item Deleting comment a) would be more justifiable than deleting comment b).
    \item Deleting comment b) would be more justifiable than deleting comment a).
    \item It would be justifiable to delete either of the comments without deleting the other.
\end{enumerate}

We grouped our sentence pairs into four "focus categories", depending on the sensitive attribute mentioned in the two sentences: Gender, Race, Religion and Mixed. For example, in the religion focus category, sentences in a pair refer to different religions. In the mixed focus category, sentences refer to different sensitive attributes, for example, $s$ might reference catholic people, while $s'$ references women. The sentence pairs shown to annotators were split, such that each annotator saw an equal number of comment pairs from each category. Details about our crowdsourcing task, instructions, payments, etc are provided in Appendix~\ref{app:data-collection}. This new dataset required approximately 4000 USD to construct (only including the payment made to the annotators and MTurk and Prolific commission).

As the demographic distributions and unalikeability coefficients among the labels collected on MTurk and Prolific were very similar to each other, we merged them into one large dataset (henceforth referred to as $\mathcal{D}_n$). Further information about platform differences can be found in \autoref{appendix:further_exploration}.

The sentence pair generation technique in \citealp{florian} aims to automatically generate good candidate pairs that many humans are likely to consider as valid fairness constraints, and our manual filtering of low-quality (semantically dissimilar) pairs is expected to exacerbate this. Correspondingly, despite the additional filtering for high unlikeability, annotations are skewed towards label option $1$ (i.e. stating that both sentences should be treated similarly). We often binarize our labels by aggregating options $2$, $3$, and $4$ to ensure each considered class contains enough labels for both learning and statistical analysis. Henceforth, we refer to option $1$ as the \textbf{\textit{Unfair} label} (i.e. sentences should be treated similarly) and the union of options $2$, $3$ and $4$ as the \textbf{\textit{Not Unfair} label} (i.e. sentences should not be treated similarly).

\paragraph{Preliminary Data Exploration} 
As our new dataset $\mathcal{D}_n$ contains the sentence pairs from $\mathcal{D}_o$ with the highest unalikeability coefficients, we expect increased variability in responses. Indeed, while $\mathcal{D}_o$ had an overall median unalikeability of 0.198 for both personal opinion and predictions of the average American opinion, $\mathcal{D}_n$ reached a median of 0.558 and 0.481 respectively. \footnote{For a fair comparison, we only considered the subset of sentence pairs in $\mathcal{D}_o$ labeled by nine different annotators.}

For conducting robust analysis with enough samples, we will binarize the race, age, and educational level characteristics of the annotators in the rest of the paper (unless otherwise stated). For the gender characteristics, we only obtained 12 participations for the Non-Binary category. In analysis involving gender, we therefore skip this category. This is a limitation of our work and should be addressed in future research.

For all demographic groups of annotators, we observe a high level of coincidence in the responses for personal opinions and predicted average American's opinion. We measured coincidence as the fraction of times annotators' personal opinion and their perception of average American opinion are the same. It ranges from 73\% for LGBTQ+ annotators to 90\% for annotators over 67 years of age; while all demographics believe their views relatively align with the average American's, such belief is strongest for people over 67 years of age and weakest for LGBTQ+ people.

\begin{figure}[h]
\centering
\includegraphics[width=0.9\columnwidth]{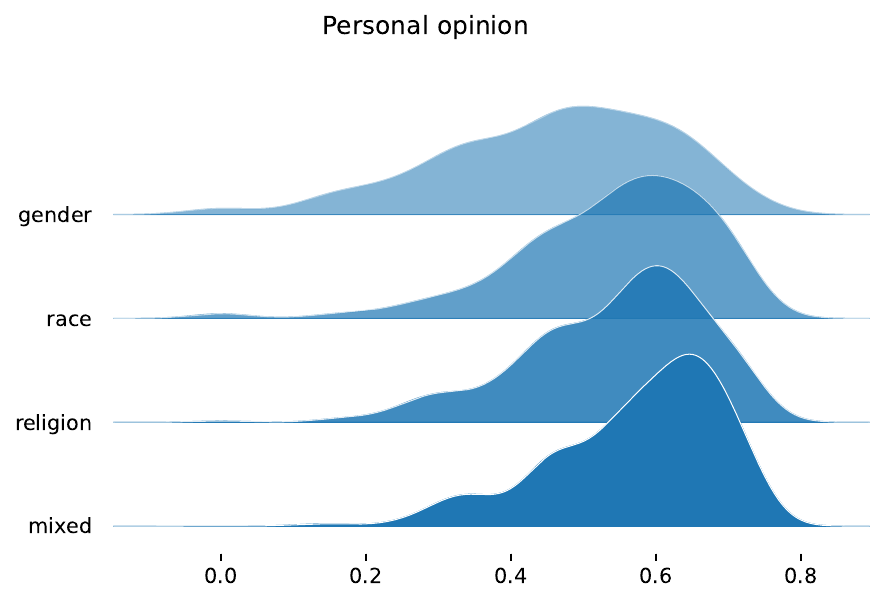}
\includegraphics[width=0.9\columnwidth]{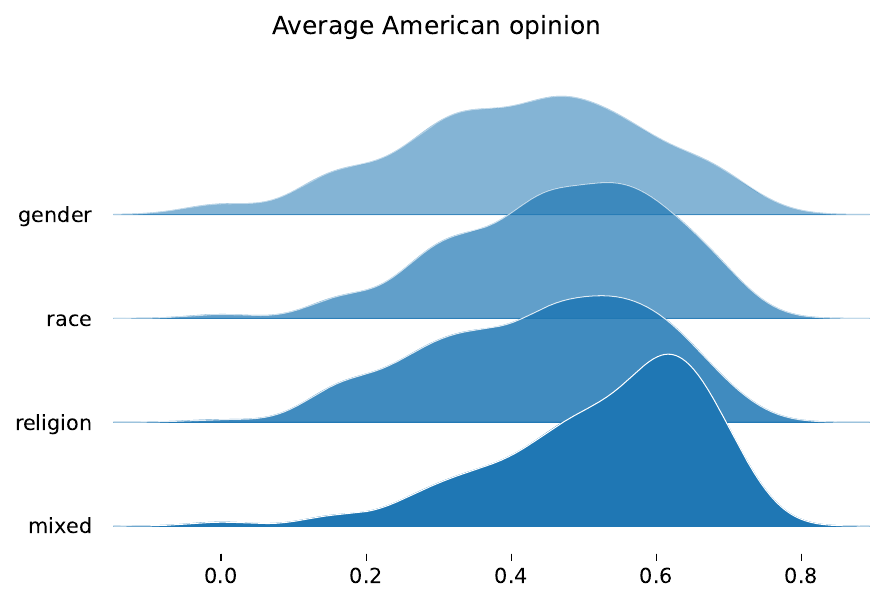}
\caption{Distribution of unalikeability coefficients in each focus category of sentence pairs. In mixed category, sentences in a pair refer to different sensitive attributes, whereas in other categories (gender, race, religion), sentences in a pair refer to different values (e.g. men, women) of the same sensitive attribute.}
\label{fig:hist_unlike}
\end{figure}

Finally, we look at the unalikeability $u_2(s,s')$ for our four focus categories of sentence pairs.
\autoref{fig:hist_unlike} shows the unalikeability coefficients for the labels of sentence pairs in each category. Sentences in the mixed focus category are noticeably more skewed towards larger unalikeability coefficients, meaning that responses are considerably more variable for such sentence pairs. This is true for both personal opinions and predictions for the average American opinion. For a more detailed statistical analysis, consider \autoref{tab:compare_unlike} in \autoref{appendix:ks-test-topics}. We also observe higher variability in annotators' personal opinions than in their perception of average Americans' opinion. \autoref{fig:unalike-personalvsavam} in \autoref{appendix:ks-test-topics} shows the variance in the two types of answers.

\begin{table*}
\begin{center}
\small
\begin{tabular}{m{0.2\linewidth} m{0.2\linewidth} m{0.2\linewidth}  m{0.2\linewidth}} \hline
 Demographic &
 Reference &
 Personal Opinion \{Odds \linebreak (\textit{P}-value)\}&
 Average American \{Odds \linebreak (\textit{P}-value)\}\\
\hline
 \textbf{Gender} & & & \\
Male & Female & 1.01 (\textit{P}=.8) & 0.97 (\textit{P}=.26)\\\hline
 \textbf{Race} & & & \\
 White & Non-White & 0.94 (\textit{P}=.21) & \textbf{0.91} (\textit{P}=.02)\\ \hline
 \textbf{Age} & & & \\
 38+ & Under 38 & \textbf{0.85} (\textit{P}<.001) & \textbf{1.27} (\textit{P}<.001)\\\hline
 \textbf{Politics} & & & \\
Independent & Democrat & \textbf{0.81} (\textit{P}<.001) & \textbf{0.68} (\textit{P}<.001)\\
Republican & Democrat & 1.10 (\textit{P}=.07) & \textbf{1.13} (\textit{P}=.004)\\\hline
 \textbf{Education} & & & \\
No High School & Bachelor & 0.56 (\textit{P}=.013) & \textbf{0.46} (\textit{P}=.001)\\
High School & Bachelor & \textbf{0.79} (\textit{P}<.001) & \textbf{0.72} (\textit{P}<.001)\\\hline
 \textbf{LGBTQ+} & & & \\
LGBTQ+ & Non-LGBTQ+ & 0.94 (\textit{P}=.35) & \textbf{1.25} (\textit{P}<.001)\\\hline
\end{tabular}
\caption{Regressions results, with labels provided by annotators as dependent variables and demographics as independent variables. 
Example: Annotators with High School education have 0.79 times the odds of providing a \textit{Not Unfair} label (i.e. sentences in a given pair should not be treated similarly), in comparison to those with Bachelor education (who are taken as reference) with \textit{P}-value$ <.001$. Results are separated for labels regarding annotators' personal opinions and their perception of the average American opinion. }
\label{tab:non-neural-1}
\end{center}
\end{table*}

\section{Differences in Fairness Preferences}\label{sec:diff}
\paragraph{Method} In this section, we study the effect of annotators' demographic identities on their own fairness judgments and their predictions of the average American's fairness judgment.  Following the standard analysis methods used in prior work (e.g.  \citealp{goyal2022your}, \citealp{kumar2021designing}, \citealp{larimore-etal-2021-reconsidering}), we predict fairness judgments based on annotators' demographic information using logistic regression. 
Because of the large class imbalance in our annotations, we focus on binarized labels, as explained in Section~\ref{sec:data}. More concretely, in a single regression we use all demographic variables $x_i^j$, where $i$ stands for gender, race, age, political stance, educational level, or the LGBTQ+ identity of annotator $j$ to predict their fairness judgments $y_{k,m}^j \in \{\text{unfair}, \text{not unfair}\}$, where $k\leq 20$ enumerates all fairness judgments by annotator $j$ and $m$ either represents their own judgment or their prediction for the average American's judgment. More details about about this analysis are in  \autoref{appendix:reg-implement}. We account for multiple hypothesis testing using Bonferroni's correction~\cite{armstrong2014use} in all reported p-values.

\paragraph{Results} Table~\ref{tab:non-neural-1} shows the per demographic odd ratios relative to a reference demographic category obtained from the logistic regression.

We observe significant effects of age, politics, and education on annotators' personal fairness judgments, and significant effects of all considered demographic variables except for gender on the predicted average American's opinion. 

We generally obtain more significant results (i.e. lower p-values) for the prediction of average American's opinion compared to annotators' personal judgments. Thus, we have stronger evidence that their demographics affect annotators' perception of the average American's judgment rather than their own judgments.

We observe that White annotators have 0.94 and 0.91 times the odds of answering \textit{Not Unfair} compared to Non-White annotators when asked about their personal fairness judgments or their predictions for the average American's judgment respectively. Only the latter effect is significant at the standard level of $p=0.05$, suggesting that White annotators are more likely to believe that the average American would prefer sentences within a pair to be treated similarly.

While we do not observe significant results for the personal opinions of Republican annotators, the odds relative to Democrats for personal and predicted average American opinion appear directionally consistent at 1.10 and 1.13. Interestingly, rather than being between Democrats and Republicans, independent annotators are more likely to believe that sentence pairs should be treated similarly (and that the average American believes the same) than supporters of both parties.

A more coherent picture emerges for education: Annotators without a High School education have 0.56 and 0.46 times the odds of answering "not unfair" compared to annotators with at least a Bachelor's education. The same tendency exists for annotators with a High School degree (odds 0.79 and 0.72). That is to say, as annotators obtain further education, both their likelihood of believing that sentences within a pair should be treated similarly, and their perception of the average American's likelihood of doing so appear to decrease.

Lastly, perhaps surprisingly, the odds are 0.85 and 1.27 respectively for age 38+ relative to the younger group. This means that younger annotators are more likely to state that they believe sentences receive similar treatment, compared to older annotators. However, they are also \textit{less} likely to think that the average American would want a pair to receive similar treatment. The same pattern can be observed for the LGBTQ+ demographic, where the respective relative odds are 0.94 and 1.25 for LGBTQ+ people. While only the effect on the perceived average American's opinion is significant in that case, it is quite interesting that both age and LGBTQ+ status show the same pattern of divergence between personal and perceived majority opinion between groups.

\subsection{Further Analysis}
\paragraph{Relation with Sensitive Attribute in Sentence Pairs}
We similarly train logistic regression models to predict annotators' fairness judgments based on which of the three focus categories (gender, race, religion) a sentence pair belongs to. 

As shown in Table~\ref{tab:regression-topic}, we find that the sensitive attribute mentioned in the sentence pairs have a significant impact on annotators' fairness judgments. Specifically, if the sensitive attribute referred in the sentence pair is gender, the odds of classifying the pair $(s,s')$ as "not unfair" are reduced by a factor 0.68 and 0.81 respectively, compared to sentence pairs in which the sensitive attribute mentioned is race. For religion compared to race, the odds are 0.98 and 0.91. In other words, sentence pairs in which sentences differ along gender are most likely to be annotated as deserving similar treatment.

\begin{table}
\begin{center}
\small
\begin{tabular}{m{3.8em} m{3.8em} m{5.5em}  m{5.5em}} \hline
 Sensitive Attribute &
 Reference &
 Personal Opinion &
 Average American \\
\hline
 Gender & Race & \textbf{0.68} (\textit{P}<.001) & \textbf{0.81} (\textit{P}<.001)\\\hline
 Religion & Race & 0.98 (\textit{P}=.61) & \textbf{0.91} (\textit{P}=.02)\\\hline
\end{tabular}
\caption{Odds ratios for predicting labels based on the sensitive attribute referenced in the sentence pair, via logistic regression.}
\label{tab:regression-topic}
\end{center}
\end{table}

\begin{table*}
\centering
\begin{tabular}{m{6em} m{6em} m{6em} m{6em} m{6em} } 
\hline
 \multirow{2}{*}{Train set} &
    \multicolumn{3}{c}{Test set} & \\ \cline{2-4}
 & Female & Male & Control & \\ [0.5ex] 
\cline{1-4}
 Female & 62.0 $\pm$ 1.8 & 60.7 $\pm$ 1.6 & \textbf{62.9 $\pm$ 1.7} & \\
 Male & \textbf{62.7 $\pm$ 1.1} & 59.8 $\pm$ 1.3 & 61.6 $\pm$ 1.1 & \\
 Control & \textbf{63.4 $\pm$ 1.3} & 60.4 $\pm$ 0.9 & 62.4 $\pm$ 1.0 & \\
\cline{1-4}
 & Non-White & White & Control & \\
\cline{1-4}
 Non-White & 55.6 $\pm$ 1.1 & 58.1 $\pm$ 1.5 & \textbf{58.3 $\pm$ 1.3} & \\
 White & 58.4 $\pm$ 1.1 & \textbf{66.5 $\pm$ 1.3} & 64.2 $\pm$ 0.6 & \\
 Control & 57.4 $\pm$ 1.2 & \textbf{64.5 $\pm$ 1.4} & 62.4 $\pm$ 1.0 & \\
\cline{1-4}
 & Under 38 & 38 + & Control & \\ 
\cline{1-4}
 Under 38 & 60.5 $\pm$ 1.7 & \textbf{62.9 $\pm$ 2.3} & 62.3 $\pm$ 2.0 & \\
 38 + & 55.0 $\pm$ 1.1 & \textbf{59.1 $\pm$ 1.8} & 56.1 $\pm$ 1.2 & \\
 Control & 60.0 $\pm$ 0.8 & \textbf{66.0 $\pm$ 1.2} & 62.4 $\pm$ 1.0 & \\
\hline
 & Democrat & Independent & Republican & Control \\ [0.5ex] 
\hline
 Democrat & 57.7 $\pm$ 1.0 & \textbf{63.9 $\pm$ 2.0} & 58.8 $\pm$ 1.4 & 60.8 $\pm$ 1.2 \\
 Independent & 57.5 $\pm$ 1.1 & \textbf{62.2 $\pm$ 1.3} & 58.1 $\pm$ 1.6 & 59.9 $\pm$ 1.3 \\
 Republican & 58.0 $\pm$ 1.0 & \textbf{63.8 $\pm$ 1.2} & 58.9 $\pm$ 1.5 & 60.8 $\pm$ 0.9 \\
 Control & 58.6 $\pm$ 1.1 & \textbf{66.9 $\pm$ 1.7}  & 60.3 $\pm$ 1.6 & 62.4 $\pm$ 1.0 \\
\hline
\end{tabular}
\caption{95\% CIs of balanced accuracy scores, obtained using 20 different random seeds for train-test splits.}
\label{tab:pairwise-classifier-comparison}
\end{table*}

\section{Effect on Downstream Models}\label{sec:effect-on-ai}
\textbf{Method} In the previous section, we provided evidence for wide differences in the fairness preferences depending on annotators' demographic identities. In this section, we study whether these differences matter for machine learning models trained to predict human fairness preferences. Such an intermediate model of human fairness preferences can be used to train fairer classifiers for a downstream task (here content moderation) \cite{florian} by creating a large pseudo-labeled set of fairness constraints from unlabeled sentence pairs.  Note that, beyond the specific settings considered in this paper, in the popular RLHF (Reinforcement Learning from Human feedback) framework, an intermediate reward model is trained from human feedback as well, which is then used to align an AI system such as a large language model~\cite{bai2022training}. It is reasonable to assume that if the intermediate model learns preferences of specific demographics, it will also impose those learned preferences in the downstream tasks. We focus on intermediate model in our setting of fairness in content moderation, but perform analysis without restricting it to a very specific task setting. We train several neural networks based on labels provided by different demographic groups separately. We hypothesize that, when tested on the data from the entirety of the population, or on the data from different demographic groups, these classifiers will learn noticeably different fairness preferences.

As $\mathcal{D}_n$ was specifically selected to only include sentence pairs with high unalikeability coefficients, which are plausibly particularly hard to classify, we found that $\mathcal{D}_n$ alone was not sufficient for training classifiers that clearly beat the random baseline. For more details, consider~\autoref{appendix:challenges-training}. Correspondingly, we decided to supplement $\mathcal{D}_n$ with all sentence pairs from $\mathcal{D}_o$. As most pairs in $\mathcal{D}_o$ only contain annotations for the predicted average American's opinion and not annotators' personal judgments, we only focus on the former. We were able to obtain data on annotator demographics, except for LGBTQ+ identity and education, from the authors of $\mathcal{D}_o$. Correspondingly, we restrict our analysis to the remaining other demographic categories (gender, race, age, politics). Similar to the previous section, we use binarized "unfair"/"not unfair" labels in the analysis.

We created subsets of $\mathcal{D}_{o+n}$, grouping by the demographic categories of annotators. For each demographic variable, we created $l+1$ subset datasets $D(\kappa)$, where $l$ is the number of categories in a demographic variable and $\kappa$ represents the value of the demographic variable or the union of these values. (e.g. the \textit{gender} variable has $\kappa=\textit{Female}$ and $\kappa=\textit{Male}$ and $\kappa=\textit{Male and Female}$. Here, $l=2$). For each demographic variable, each of the first $l$ subsets thus contains data annotated solely by participants who belong to the $l^{th}$ category in that demographic variable. The $l+1^{th}$ dataset contained all datapoints, regardless of the annotators' demographics, which we denote as the Control dataset. For each of theses subsets, we fitted pre-trained multiheaded BERT based classifiers~\cite{kenton2019bert} building upon the code by ~\citealp{florian}. Instead of aggregating individual annotations (e.g. into majority vote) in the training set, we used all the annotations for training the classifiers~\cite{wei2023aggregate}. In \autoref{appendix:challenges-training}, we discuss that the accuracy of classifiers trained with only majority vote was worse and that was the reason for using all annotations for training. 

We trained and tested multiple classifiers $\phi^{\kappa}_i$ for each subset $D({\kappa})$, where $i \in [1,...,20]$ corresponds to one of 20 different random seeds for the creation of the train, validation (10\%), and test sets (10\%) $D_i^{train}(\kappa)$,$D_i^{val}(\kappa)$,$D_i^{test}(\kappa)$. We provide further details about training the classifiers and hyperparameter tuning in \autoref{appendix:training_details}.
 Then, as done by  \citealp{al2020identifying} \citealp{binns2017like}, and \citealp{akhtar2020modeling}, we compared the set of models $\phi^{\kappa}_i$ and $\phi^{\kappa'}$, trained with data labeled by annotators from different demographic categories $\kappa$ and $\kappa'$ respectively (e.g. we compare the models trained with data annotated by Democrat participants with the models trained with data annotated by Republican and Independent participants separately). Each model $\phi^{\kappa}_i$ is evaluated on all $D_i^{test}(\kappa')$ for $\kappa'$ in the same demographic group (e.g. gender). Before that, in order to account for label imbalances, we optimized the classification threshold used by $\phi^{\kappa}_i$ on $D_i^{val}(\kappa')$ to maximize the models' balanced accuracy~\cite{balanced-accuracy} on the demographic $\kappa'$.

\begin{table*}[h!]
\begin{tabular}{m{4.6em} m{6em} m{3.2em} m{3.6em} m{5.5em} m{5.5em} m{2.5em}}
 \toprule
Gender & Race & Age & Politics & Ensemble  & Control & Size\\
 \midrule

Female & None-White & $<$ 38 & Dem. & 57.1 $\pm$ 2.9 & 55.7 $\pm$ 2.7 & 32 \\
& &  & Indep. & 50.4 $\pm$ 2.8 & 53.2 $\pm$ 3.6 & 19\\
& &  & Rep. & 57.2 $\pm$ 3.6 & 52.0 $\pm$ 5.1 & 7\\

& & $\geq$ 38 & Dem. & 52.7 $\pm$ 2.4 & 52.5 $\pm$ 2.2 & 194\\
&  &  & Indep. & 54.1 $\pm$ 2.2 & 52.9 $\pm$ 2.6 & 140\\
&  &  & Rep. & \textbf{54.8 $\pm$ 4.9} & 44.7 $\pm$ 5.2 & 5\\
 \midrule
& White & $<$ 38 & Dem. & \textbf{50.1 $\pm$ 1.7} & 47.4 $\pm$ 1.7 & 79\\
&  &  & Indep. & 57.7 $\pm$ 2.7 & 56.5 $\pm$ 3.4 & 38\\
&  &  & Rep. & 55.5 $\pm$ 2.5 & 57.0 $\pm$ 3.8 & 23\\

&  & $\geq$ 38 & Dem. & \textbf{64.0 $\pm$ 1.1} & 60.6 $\pm$ 1.6 & 290\\
&  &  & Indep. & \textbf{69.6 $\pm$ 1.7} & 63.2 $\pm$ 1.5 & 198\\
&  &  & Rep. & 57.4 $\pm$ 1.7 & 56.6 $\pm$ 1.9 & 150\\
 \midrule
Male & Non-White & $<$ 38 & Dem. & 49.6 $\pm$ 1.7 & 50.8 $\pm$ 2.1 & 56\\
& &  & Indep. & 48.2 $\pm$ 3.2 & 49.2 $\pm$ 4.5 & 30\\
& &  & Rep. & \textbf{51.5 $\pm$ 3.2} & 44.9 $\pm$ 4.5 & 10\\
& & $\geq$ 38 & Dem. & \textbf{59.8 $\pm$ 1.5} & 56.5 $\pm$ 1.5 & 377\\
&  &  & Indep. & \textbf{66.0 $\pm$ 2.5} & 59.4 $\pm$ 2.5 & 104\\
&  &  & Rep. & 62.2 $\pm$ 3.2 & 58.4 $\pm$ 4.0 & 36\\
 \midrule
& White & $<$ 38 & Dem. & 51.2 $\pm$ 2.3 & 51.1 $\pm$ 2.3 & 75\\
&  &  & Indep. & 49.0 $\pm$ 4.8 & 45.0 $\pm$ 5.3 & 34\\
&  &  & Rep. & 51.9 $\pm$ 2.9 & 53.3 $\pm$ 2.8 & 35\\
&  & $\geq$ 38 & Dem. & \textbf{60.4 $\pm$ 1.4} & 56.1 $\pm$ 1.3 & 350\\
&  &  & Indep. & \textbf{67.7 $\pm$ 2.0} & 63.7 $\pm$ 2.1 & 141\\
&  &  & Rep. & 63.7 $\pm$ 1.6 & 62.5 $\pm$ 1.4 & 175\\
 \bottomrule
\end{tabular}
\caption{95\% CIs for the balanced accuracy scores for the ensemble classifier and the Control model, per demographic intersection, along with the number of annotators belonging to each intersection. Significant (at $p=0.05$) improvements of the ensembling approach according to a t-test are in \textbf{bold}.}
\label{tab:intersectional}
\end{table*}

\paragraph{Results}
Table~\ref{tab:pairwise-classifier-comparison} shows that the performance of different classifiers vary noticeably depending on the training and test data used. However, there is no clear trend towards classifiers performing best for the demographic they were trained on. With few exceptions, the test set containing data annotated by a single certain demographic $\kappa \in \delta$ in a group of demographics $\delta$ tends to show higher levels of balanced accuracy~\footnote{\autoref{fig:specificity} in the Appendix also shows specificity and sensitivity scores for different models.} across all models $\phi^{\kappa'}, \kappa' \in \delta$. This is consistent with similar observations made by \citealp{binns2017like} in their search for differences in judgments of offensiveness across genders, where they found that all classifiers perform better when evaluated on data annotated by men rather than women. It appears that classifiers are able to replicate preferences of annotators in some demographic groups (women, White people, people over 38 years of age and politically independent people) better than those of other demographic groups, potentially regardless of the training data. The reasons for this are not entirely clear;  different data set sizes and levels of disagreement within group may be contributing factors. We compare some of these factors across demographics in \autoref{appendix:similaries-groups}, and leave an in-depth investigation for future work.

Despite the lack of a clear directional effect of the training demographics,  pairwise Kolmogorov-Smirnov tests~\cite{ref1} comparing the population of trained models $\phi^{\kappa}$ $\phi^{\kappa'}$ for different demographics $\kappa$ and $\kappa'$ in terms of balanced accuracy on the control test set yield statistically significant differences for gender (\textit{P}<$.05$), race (\textit{P}<$.001$), and age (\textit{P}<$.001$). This suggests, that the differences in fairness judgments based on annotator demographics observed in Section~\ref{sec:diff} do affect downstream machine learning models, albeit perhaps not in a clearly interpretable manner.

\section{Ensemble Classifier}\label{sec:ensemble}
Previous work has observed disproportionate performance drops for demographic intersections, for example black women, compared to both black people or women \cite{may2019measuring,tan2019assessing}.
We investigated how our Control model that gives equal consideration all data points, regardless of the demographics of annotators, performs when tested on the labels provided by different demographic intersections (e.g. black women). We hypothesize that demographic intersections with a smaller number of annotators will tend to have worse performance because they are underrepresented in the Control model's training data. For example, in $\mathcal{D}_{o+n}$, we have annotations from 377 (14.4\%) Non-White, over 38 years of age, Democrat men, but only 5 (0.02\%) Non-White, over 38 years of age, Republican women. We evaluated each of the 20 random seed versions of the Control model using only test labels provided by a specific demographic intersection. We then regressed the models' balanced accuracies on the number of annotators per intersection, finding a positive correlation (with \textit{P}$ =.015$).

To address this, we use an ensemble classifier $\bar{\phi}$ that aggregates the previously trained classifiers $\phi^\kappa$ via majority voting. We hypothesize that by giving the same weight to each demographic, such an ensemble model will provide better results for a larger number of demographic intersections than our Control model, which gives the same weight to all labels.
We compare the predictions $\bar{\phi}(s,s')$ with each intersection's majority vote annotations and calculate balanced accuracy. 

\autoref{tab:intersectional} shows the balanced accuracy scores obtained for our ensemble classifier and the Control model per demographic intersection. The ensemble classifier $\bar{\phi}$ provides significantly better (p<0.05) results for 9 out of 24 demographic intersections while the Control model did not provide significantly higher scores for any demographic intersection (and non-significantly better results for 5 out of 24 intersections). For coarse demographics categories (\autoref{tab:non-intersectional}), we observe a significant improvement in the performance of 8 out of 9 demographics. 

\begin{table}
\begin{tabular}{m{7em}m{4em}m{4em}} 
\hline
Demographic & Ensemble & Control  \\
\hline
Gender & & \\
\hspace{3mm} Female & \textbf{66.1$\pm$1.0} & 62.2$\pm$1.0\\ 
\hspace{3mm} Male & \textbf{64.3$\pm$0.7} & 60.2$\pm$0.8\\  \hline
Race & & \\
\hspace{3mm} Non-White & \textbf{60.5$\pm$1.1} & 56.5$\pm$1.3\\ 
\hspace{3mm} White & \textbf{68.4$\pm$0.8} & 63.0$\pm$0.9\\ \hline
Age & &  \\
\hspace{3mm} Under 38 & 55.6$\pm$1.9 & 55.8$\pm$2.0\\ 
\hspace{3mm} $\geq$ 38 & \textbf{66.0$\pm$0.6} & 61.3$\pm$0.7\\ \hline
Politics & & \\
\hspace{3mm} Democrat & \textbf{61.4$\pm$1.0} & 57.7$\pm$0.9\\ 
\hspace{3mm} Independent & \textbf{70.6$\pm$1.2} & 66.0$\pm$1.4\\
\hspace{3mm} Republican & \textbf{63.1$\pm$1.4} & 61.7$\pm$1.6 \\
\hline
\end{tabular}
\caption{Balanced accuracy scores (with 95\% CIs) for the ensemble classifier and the Control model, per demographic variable. Significant (at $p=0.05$) improvements of the ensembling approach according to a t-test are in \textbf{bold}.}
\label{tab:non-intersectional}
\end{table}

Interestingly, while our approach provides much better scores for our smallest demographic intersection (Non-White, older, Republican women), improving its balanced accuracy score by more than 10 percentage points, we also observe a performance improvement for White, older, Democrat women (our largest demographic intersection). Our analysis for fairness thus complements the analysis from prior work on dealing with disagreements~\cite{davani-etal-2022-dealing,gordon2022jury}. 

\section{Conclusions}
Human feedback is often used to improve artificial intelligence systems by learning to approximate human preferences such as fairness constraints. Via a crowdsourcing study, we investigated the role of different demographic factors on human annotators' fairness preferences. We focused on a content moderation task as example, where annotators were asked to compare how two sentences should be treated in comparison to one another. We found that demographic variables (age, politics, education and LGBTQ+ identity) play a significant role in the annotations. We observed this effect for annotators' personal opinions as well as their perception of the average American opinion. Further, we found that differences in the fairness preferences also influence the performance of models trained to predict these preferences. For example, models trained on annotations from different demographic categories, when evaluated on a control test set, gave statistically different balanced accuracy scores. Our results highlight the ethical implications of using human feedback to align AI systems with fairness preferences since the preferences of some demographic groups may be more dominantly reflected in the downstream systems. Finally, we observed that an ensemble classifier that uses the majority vote of models trained on annotations from different demographic categories can improve accuracy scores for a number of demographic intersections.

\section{Limitations}\label{sec:limitations}
Since crowdsourcing platforms like MTurk and Prolific are frequently used for collecting data for training machine learning classifiers, we focus on crowd worker judgements. We acknowledge that the opinions of crowd workers are not necessarily representative of the wider population (within the US or beyond). Therefore, conclusions about opinions of different demographics should be interpreted with this difference in mind and only carefully be generalized to other settings. 

Our survey did not include data on several kinds of personal attributes such as disability status. We also bring attention to the lack of data in our survey from non-binary gender group. Further, we also binarized several attributes like age, race and educational status in our analysis due to data constraints, but future research should examine this aspect further. 

Our survey was formulated in English and the workers were expected to know English; the results can therefore not be  generalized for other languages and cultures, without further data. We also note that our survey focused on sentence pairs with particularly contentious fairness judgements. Further, our data focuses on content moderation domain only. 

Our experiments on determining the effect of demographic differences in annotations on downstream models focused on the models that were trained to predict fairness judgments. As explained in the paper, this is a reasonable assumption for task-agnostic analysis (within the broader domain of content moderation). In future work, it will be interesting to investigate the effects on downstream models trained to respect the learnt fairness judgments. 

The term "average-American" used in our survey is not very specific (compared to for e.g. "other annotators on MTurk, located in the US and with such and such qualifications"). The intuition behind asking two separate questions (personal opinion and perception of opinion of average-American) was to elicit clear information, even if the participants believed that their own opinions would differ from what they might assume to be a more commonly held opinions in the US. But we have not explored in this work how a different term could affect the complexity of the survey, the experience of participants and the results of the experiment.

Finally, we note that the positive correlation between performance and size of the demographic intersections does not fully disappear with ensemble approach and more generally, differences in performance across groups continue to exist. Therefore, future research should continue to explore solutions to dealing with disagreements.

We also remind the reader, that because of data availability issues, the results in Sections~\ref{sec:effect-on-ai} and \ref{sec:ensemble} pertain to annotators' perception of the average American's fairness judgments, rather than their personal fairness judgments.

\section{Acknowledgments}
This work was partially supported by the Oxford Martin School's programme on Ethical Web and Data Architectures in the Age of AI (EWADA). The emojis in \autoref{fig:approach} were designed by OpenMoji – the open-source emoji and icon project (license: CC BY-SA 4.0).

\bibliography{anthology,custom}
\bibliographystyle{acl_natbib}
\clearpage
\appendix
\noindent \textit{\textcolor{red}{Warning: Following pages contain examples of language that may be offensive or disturbing.}}
\section{Appendix A: Data Collection Details}
\label{sec:appendix}

\subsection{Selection of Sentence Pairs for $\mathcal{D}_n$}\label{appendix:selection}

The first task in our data collection process was to determine an appropriate size for dataset $\mathcal{D}_n$; the amount of sentence pairs $n_{n}$ to be included and the amount of labels to be collected per sentence pair. We preliminarily used dataset $\mathcal{D}_o$ to analyze the effects of decreasing the dataset size on the significancy of results, to find a lower threshold on the amount of data necessary to obtain statistically significant results in regards to the differences between the models trained with data from opposite demographic categories. Initially, we used the entirety of $\mathcal{D}_o$ and ten different train/validation/test splits in our data to create ten iterations of the best-performing BERT or RoBERTa models $\phi$ when trained solely on data annotated by demographic category $\kappa$, $\phi_i^{\kappa}$, $i \in [1, ..., 10]$, $\kappa \in \{female, male, Non-White, White\}$. Then, each iteration $\phi^\kappa_i$ was evaluated once for every category $\kappa'$ from its own demographic variable $\delta$ using as test set the labels provided by annotators from such demographic category $\kappa'$. After the creation of 10 iterations of each model, we used a Student's t-test~\cite{student1908probable} to observe whether differences in the labeling patterns of different demographic categories were significant. After observing significant differences for models trained on all $\mathcal{D}_o$, we gradually reduced to number of sentence pairs with which we trained and tested the BERT and RoBERTa models and repeated such procedure until the differences in the perception of individual fairness across demographics were no longer significant. We found that the classifiers built with a dataset conformed by 1421 sentence pairs and at least 4 participations per sentence pair were still able to recognize the differences between demographic categories $\kappa$ significantly ($\alpha = 0.05$). Given the allocated budget, we were able to collect data on 1,500 sentence pairs, having at least 10 worker labels per pair. We opted for increasing the number of labels per sentence pair to have a larger and more diverse pool of annotators' demographics per pair and therefore obtaining more robust responses when filtering by demographic traits.

To select the sentence pairs to be included in $\mathcal{D}_n$, we analyzed dataset $\mathcal{D}_o$ to give priority to sentence pairs with a higher level of disagreement among participants, as to explore if the demographic characteristics of each worker influenced such differences. We handpicked the sentences with the highest response variability and manually corrected the sentences that either had spelling or grammatical errors (i.e. \textit{why should not the foster father be permitted to adopt the infant that has bonded with her?}) or that were factually incorrect. On the other hand, certain sentence pairs were not considered because of the lack of relationship between sentences $a$ and $b$ or because of incoherence in sentence $b$. As stated before, some sentence pairs provided by \citealp{florian} rely on word replacement, taking sentence $a$ as a base and creating sentence $b$ by replacing the words referencing a demographic. The authors note that methodology oftentimes returns incoherent phrases (i.e. replacing "The President's White House" with "The President's Black House" would not be correct). On the other hand, we also observe pairs that do not show much similarity within their sentences and lack a strong relation between them. Some examples of sentence pairs that were disregarded for such reasons, independently of the level of variability in their labels, can be seen in \autoref{tab:examples-not-included}.

\begin{table*}[h]
\centering
\begin{tabular}{m{19em} m{19em}} 
 \toprule
 Sentence $s$ & Sentence $s'$\\
 \midrule
 According to Neil Black, no crimes were ever committed \textit{... [continued]} & According to Neil Homosexual, homosexuals were not persecuted \textit{... [continued]}\\
 \\ 
 Donald will be impeached (or more likely, resign before being impeached) and I'm glad I don't live in your reductive, black and white world. & President Trump is an atheist.\\ 
 \\
 I'm thinking I must have some of that yellow or white fat because I'm pretty sure I'm not Greenland Inuit!  Argh!  lol \textit{... [continued]} & i'm gay. gay  \\
 \\
 These people do not understand that Roman Catholic religious ed is only for kids \textit{... [continued]} & This woman is mentally ill \\ [1ex] 
 \bottomrule
\end{tabular}
\caption{Examples of sentences with high unalikeability coefficients $u_2(s,s')$ which were not considered for the new dataset $\mathcal{D}_n$. Please note, that the second shown sentence from column $s$ is a sentence that supposedly references Black people, which is inaccurate.}
\label{tab:examples-not-included}
\end{table*}

Regarding the sensitive attribute mentioned in each sentence pair, we ensured that all participants were shown an identical amount of sentence pairs from each focuc category- Gender, Race, Religion, and Mixed. \autoref{tab:examples-by-focus} provides sample sentences of such sentence pairs. We did not add a section on sentences that solely refer to the LGBTQ+ community because dataset $\mathcal{D}_o$ only contained around 120 pairs of this kind and we soon realized that the sentences related to such topic had a lower qualitative quality than the rest. Also, we did not include pairs where the first sentence regarded gender and the second sentence regarded a different demographic trait: we found that most words that refer to people are gendered (i.e. \textit{lady, guy, girl, son, etc.}) and changing these to a word that references a person's religion, race or sexuality often resulted in extremely offensive sentences.

\begin{table*}[h]
\centering
\begin{tabular}{m{16.5em} m{16.5em} m{4em}} 
 \toprule
 Sentence $s$ & Sentence $s'$ & Sensitive Attribute\\
 \midrule
Lots of immigrants live in Scandinavia.   Plus they elect female leaders. Want to try it? &  Lots of immigrants live in Scandinavia.   Plus they elect male leaders. Want to try it? & Gender\\
 \\ 
White supremacists have just as many rights afforded them as anyone else. &  Latinos Terrorists have just as many rights afforded them as anyone else. &  Race \\
\\
What town do you live in and how many Muslims live there? & 
What town do you live in and how many Lutherans live there? & Religion\\
 \\
Buddhists three poisons:  anger, greed, and pride. & Homosexual people three poisons:  anger, greed, and pride.  & Mixed \\[1ex] 
 \bottomrule
\end{tabular}
\caption{Examples of sentence pairs along with the sensitive attribute mentioned in the respective pair.}
\label{tab:examples-by-focus}
\end{table*}

\subsection{Data Collection Process}\label{app:data-collection}

For our collection task, we recruited workers from Amazon Mechanical Turk and Prolific to later redirect them to our survey which was designed and hosted on Qualtrics. To ensure high-quality data and a speedy collection process, we added a Captcha at the beginning of the survey, enabled the flagging of participants who answered the survey at an unusually high speed, added a function to prevent workers from responding to the survey more than one time, and only allowed users to enter the survey if they were being redirected from either MTurk or Prolific. Although Google's technology reCaptchav3 could also be implemented in Qualtrics for bot detection purposes, we refrained from its use because of its lack of GDPR compliance.

We required all participants to have a minimum amount of completed surveys, a minimum acceptance rate and be located in the US. In the case of Mechanical Turk, we requested workers to have completed at least 5,000 previous tasks, with an acceptance rate of 95\% or higher, as \citealp{florian} did in their data collection process. On the other hand, for Prolific we required a minimum of 1000 previously completed tasks and a 97\% acceptance rate. We decreased the number of previously completed tasks because according to Prolific, if we were to make our survey available only to those who had completed over 5,000 tasks, we would only be able to access a pool of over 300 participants\footnote{Information provided at the moment of creating a new survey via Prolific.com}. Nonetheless, to compensate for this, we increased the threshold for the acceptance rate.

Prior to their participation, participants were informed about the existence of offensive content in the survey and assured that they could withdraw from participating at any given moment, but that completion would be required to receive monetary compensation for their work. We also provided additional information on the task, the type of data that would be collected from them, survey examples, and contact information in case of concerns. This disclaimer was originally presented to the ETH Zurich Ethics Committee in May 2023 and accepted under proposal number EK 2023-N-133-A.

While data obtained through Prolific preserved higher quality levels, Mechanical Turk presented many more submissions from what seemed bots or other workers that did not put effort in the tasks. On the first iteration of the data collection task, MTurk only delivered 140 acceptable responses out of 500 participations. We were able to identify human-like responses by requesting a written explanation to one of the previously provided answers. Moving forward, to avoid such behavior we created a new MTurk Qualification in which workers had to correctly respond to two hand-crafted multiple-choice questions to gain access to the rest of the survey. Unfortunately, upon the installment of such Qualification, the speed at which the survey was answered decreased significantly. While the first 500 submissions -regardless of their quality or approval rate- were collected in a lapse of 4 hours, we required approximately 3 weeks to collect 203 more participations once the Qualification was installed. Such a filter increased the quality of the responses drastically, raising to a 65\% approval rate instead of the original 28\%. Nonetheless, having one-third of all HITs being rejected was still an elevated rate of rejection, we opted to remove the Qualification once again for the sake of time efficiency. After doing so, we observed data quality and speed of response similar to the first released batch. Other authors have researched on the shift in the data quality provided by MTurk. While \citealp{kennedy2020shape} hypothesize on users recurring to fraudulent access to studies to improve the quality of their responses, \citealp{moss2018after} notes that poor quality responses often originate from repeated geolocations, and found 60 locations where workers were submitting HITs tied to server farms.

After each collection batch, all cases where the attention checks were failed, where the survey was responded to under 6 minutes\footnote{The median completion time rounded to 13 minutes.}, or where the written explanation did not seem coherent were manually inspected. Cases where the submission did not meet our quality standards were discarded.

Participants were provided with a pair of relatively similar sentences and then given the following queries:

\begin{itemize}
    \item Without further context, would you consider it unfair for a context moderator to delete one of the above comments but not the other?
    \begin{itemize}
        \item It would be unfair to delete one of the comments, but not the other.
        \item Deleting comment a) would be more justifiable than deleting comment b).
        \item Deleting comment b) would be more justifiable than deleting comment a).
        \item It would be justifiable to delete either of the comments without deleting the other.
    \end{itemize}
    \item What do you expect the average American to answer to the previous question?
        \begin{itemize}
        \item It would be unfair to delete one of the comments, but not the other.
        \item Deleting comment a) would be more justifiable than deleting comment b).
        \item Deleting comment b) would be more justifiable than deleting comment a).
        \item It would be justifiable to delete either of the comments without deleting the other.
    \end{itemize}
\end{itemize}

The previously shown format was taken from \citealp{florian}, which they used for their entire crowdsourcing process. However, after a preliminary release of the survey, we came to believe that the original response options were causing confusion among workers. For this reason, we modified them to a simpler version and rearranged the order in which they are shown, as previously presented. The original options read as follows:

\begin{itemize}
    \item Without further context, would you consider it unfair for a context moderator to delete one of the above comments but not the other?
    \begin{itemize}
        \item  It would be unfair for one of the two comments but not the other to be deleted (Both comments ought to be treated the same way, no matter what)
        \item  If comment a) was deleted, comment b) should be deleted as well, but not the other way around. (Deleting comment b) would be more justifiable)
        \item  If comment b) was deleted, comment a) should be deleted as well, but not the other way around. (Deleting comment a) would be more justifiable)
        \item  It would be acceptable to delete either of the comments without deleting the other. (Treating the comments differently either way would be fine)
    \end{itemize}
\end{itemize}

\paragraph{Payment} Independently of the platform used to hire workers, we compensate our participants with 2.40 USD per survey, which yields approximately 12 USD per hour, given that the median completion time rounded to 13 minutes. This exceeds the USA federal minimum wage of 7.25 USD per hour and is only exceeded by the state minimum salary of 13 states. To ensure the fairness of such compensation, we ran a preliminary smaller survey in which we analyzed the time required to respond to the survey and adjusted our compensation accordingly.

\paragraph{Analysis of Data Collected With and Without Timer} A timer of 12 minutes was set during the first batch of our data collection task on MTurk which provided 140 approved participations. While the survey had a median completion time of 13 minutes in Prolific, some workers did complain about having too little time to complete the survey on MTurk. We removed the timer in susequent batches. We compare this batch to the rest of the data collected via MTurk to assess if there were significant differences and decide whether this data could still be useful for our research or not.

As shown in \autoref{tab:timer}, the raw responses for both batches are quite similar for both the answers regarding personal opinions and the answers regarding average American opinions.

\begin{table*}[h]
\centering
\begin{tabular}{m{8em} m{5em} m{5em} m{5em} m{5em}} 
 \toprule
 \multirow{2}{*}{Response} &
    \multicolumn{2}{c}{Personal opinion} &
    \multicolumn{2}{c}{Average American} \\
 & Timer & No Timer & Timer & No Timer\\ [0.5ex] 
 \midrule
 Unfair & 57.23& 55.91 & 62.52 & 63.17\\ 
 A worse & 14.06 & 13.39 & 14.10 & 10.42\\ 
 B worse & 15.14 & 16.79 & 12.45 & 12.49\\ 
 Either & 13.56 & 13.91 & 10.94 & 13.92\\  [1ex] 
 \bottomrule
\end{tabular}
\caption{Distribution of raw responses, when the MTurk batch is set with a timer and when it is not. Amounts are presented in percentages. }
\label{tab:timer}
\end{table*}

Furthermore, we analyzed the level of disagreement per sentence pair in both datasets, via the unalikeability coefficient \cite{kader2007variability}. The timed dataset appeared to have much lower variability per response than the dataset without a timer, however this was caused by the size difference in both datasets. While there exist 1,500 sentence pairs in our dataset, in the batch collected with a timer 640 sentence pairs had one or fewer responses and 1,040 sentence pairs had two or fewer responses, which caused the low variability count to rise. In comparison, only 20 sentence pairs had one or fewer responses and 100 sentence pairs had two or fewer responses in the dataset without a timer, which naturally raised the unalikeability coefficient of answers per sentence pair. For this reason, we compared the level of disagreement in both datasets by conducting a pairwise Kolmogorov-Smirnov test on the unalikeability values of both batches, by only taking into consideration the data from sentence pairs that had exactly 2 participations in the batch with timer and the data from pairs that had 2 participations in the batch without timer; we continued by comparing the data corresponding to sentence pairs with 3 participations and so on. There is no statistically significant evidence of both batches having different unalikeability distributions except for the responses on the average American opinions where each dataset had either 2 or 5 participations. All variability differences between responses regarding personal opinions are not significant. \autoref{tab:timer-ks} shows the obtained p-values and statistics for all Kolmogorov-Smirnov tests.

\begin{table*}[h]
\centering
\begin{tabular}{m{12em} m{9em} m{9em}} 
 \toprule
Number of responses & Personal opinion & Average American\\
 \midrule
 2 & 0.6316 (0.09) & \textbf{0.0049 (0.21)}\\ 
 3 & 0.1490 (0.08) & 0.5322 (0.06)\\ 
 4 & 0.3469 (0.13) & 0.9472 (0.07)\\ 
 5 & 0.0917 (0.28) & \textbf{0.0025 (0.41)}\\  [1ex] 
 \bottomrule
\end{tabular}
\caption{Results from performing a Kolmogorov Smirnov test comparing the distributions of unalikeability coefficients between the timer dataset and the dataset without a timer, with different number of answers per sentence pair. Presented are the p-values of the test along with its corresponding statistic.}
\label{tab:timer-ks}
\end{table*}

Finally, given that some of the experiments performed in our work use the majority votes per sentence pair as labels, we analyze how such labels change when adding or removing the batch with a timer from the rest of the MTurk dataset. For the responses on the average American's opinions, 55 out of 1500 majority votes change. In these 55 cases, the answers provided by the timer dataset represent on average 47\% of the responses per sentence pair. In the case of the answers on personal opinions, 62 out of 1500 majority votes change. Here, the answers provided by the timer dataset represent on average 43\% of the responses per sentence pair. This effect is even smaller when we also take into account the data obtained from Prolific. In this case, we have 11 and 16 majority vote changes for the answers on personal opinions and the answers on the average American opinions respectively.

\section{Appendix B: Further Data Exploration}\label{appendix:further_exploration}

\begin{figure}[h]
\centering
\includegraphics[width=0.9\columnwidth]{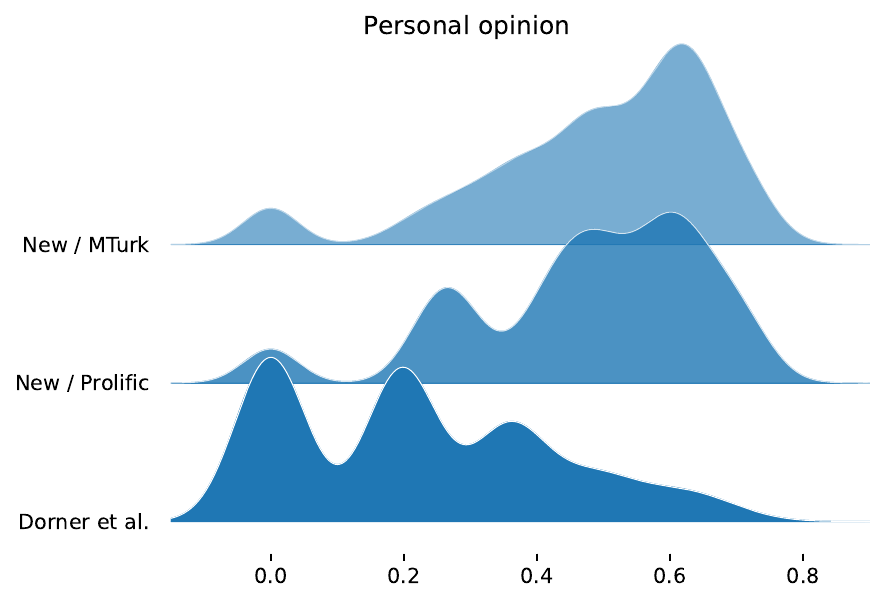}
\includegraphics[width=0.9\columnwidth]{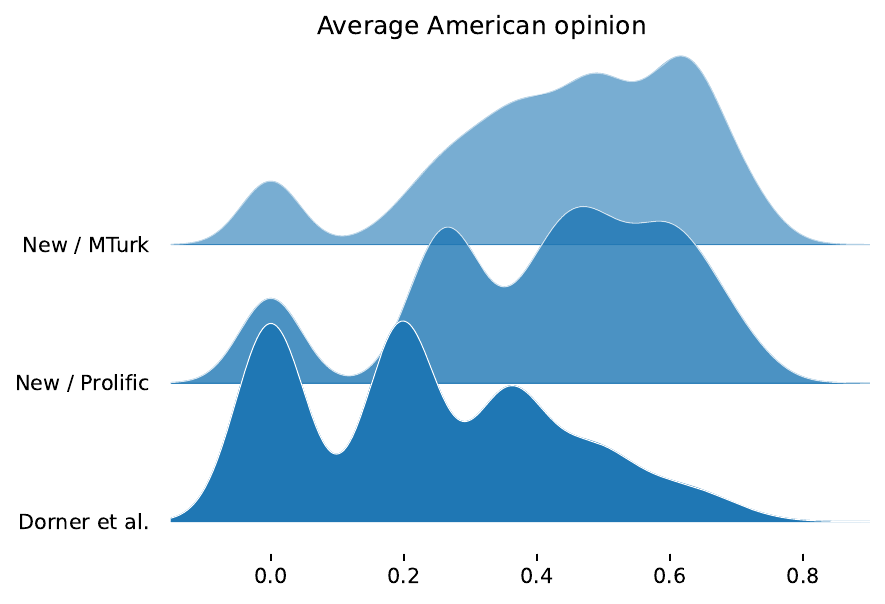}
\caption{Distribution of unalikeability coefficients per dataset.}
\label{fig:unalike-dataset}
\end{figure}

We successfully obtained higher label variability by selecting the sentence pairs that deemed higher unalikeability coefficients in $\mathcal{D}_o$. Figure \ref{fig:unalike-dataset} shows the count of unalikeability coefficients for the data collected through MTurk, Prolific and $\mathcal{D}_o$. In the figure we only include the sentence pairs from $\mathcal{D}_o$ that have 9 labels, for adding sentence pairs with less responses would naturally skew the graph further left.

Regarding the demographic distribution of annotators,  per Table \ref{tab:summary_stats}, we observe quite different distributions for age, politics, education, and LGBTQ+ identity in Prolific's and MTurk's populations. Prolific's sample is less conservative, less educated, and more sexually diverse than MTurk's sample. Our sample's demographic distribution does not align with that of the US population, which was expected; according to \citealp{levay2016demographic}, annotators on MTurk are on average 20 years younger than the ANES sample \cite{anes2021}, with a larger proportion of Democrats, which results in an under-representation of the Republican population. Finally, Levay et al. report a more educated population than the ANES sample, which can also be observed in our dataset. We do not observe high correlations between demographic traits in $\mathcal{D}_n$ -the union of the data obtained from MTurk and Prolific. However, the female population is older than the male population\footnote{60\% of women and 50\% of men had over 38 years.} and shows a higher level of sexual diversity\footnote{20\% of women and 13\% of men identified as LGBTQ+.}. Also, there is a much higher proportion of young, Non-White people than young White people\footnote{61\% of Non-White people and 40\% of White people had 37 years or less.}. On the other hand, the demographic distribution of $\mathcal{D}_{o+n}$ is shown in Table \ref{tab:summary_stats2}. We observe a very similar distribution as the one presented for $\mathcal{D}_n$, with the distinction that we cannot measure for the education and LGBTQ+ identity variables. We observe a slight shift in the race variable, with an increment in the Black people category and a decrement in the White people category. On the other hand, the percentage of annotators under 38 years went from nearly 45\% to 54\%.

\begin{table}[h]
\small
\begin{tabular}{m{8em} m{3.4em} m{3.4em} m{3.4em} }
\toprule
Demographics & Prolific & MTurk & Total\\
\midrule
Gender\\
\hspace{3mm}Female & 51.6 & 48.0 & 49.8 \\
\hspace{3mm}Male & 46.8 & 51.2  & 49.0 \\
\hspace{3mm}Non-Binary & 1.6 & 0.8 & 1.2 \\
Race\\
\hspace{3mm}Asian & 7.8 & 7.0  & 7.4 \\
\hspace{3mm}Black & 11.0  & 7.8  & 9.4 \\
\hspace{3mm}Hispanic & 6.2  & 6.8  & 6.5 \\
\hspace{3mm}White & 71.4  & 76.2  & 73.8 \\
\hspace{3mm}Other & 3.6  & 2.2  &  2.9\\
Age\\
\hspace{3mm}18 - 27 & 16.0  & 10.4  & 13.2 \\
\hspace{3mm}28 - 37 & 28.8  & 34.6  & 31.7 \\
\hspace{3mm}38 - 47 & 24.6  & 29.0  & 26.8 \\
\hspace{3mm}48 - 57 & 17.4  & 14.6  & 16.0 \\
\hspace{3mm}58 - 67 & 9.0  & 8.8  & 8.9 \\
\hspace{3mm}Over 67 & 4.2  & 2.6  & 3.4 \\
Political Stance\\
\hspace{3mm}Democrat & 52.6  & 53.8  & 53.2 \\
\hspace{3mm}Republican & 16.2  & 24.8  & 20.5 \\  
\hspace{3mm}Independent & 31.2  & 21.4  & 26.3 \\
Educational Level\\
\hspace{3mm}No High School & 1.0  & 0.4  & 0.7 \\
\hspace{3mm}High School & 44.4  & 33.0  & 38.7 \\
\hspace{3mm}Bachelor or more & 54.6  & 66.6  & 60.6 \\
LGBTQ+ Identity\\
\hspace{3mm}LGBTQ+ & 20.6  & 14.8  & 17.7 \\
\hspace{3mm}Non-LGBTQ+ & 79.4  & 85.2  & 82.3 \\
\bottomrule
\end{tabular}
\caption{Demographic distribution of annotators in $\mathcal{D}_n$ separated by platform. Results are presented in percentages.}
\label{tab:summary_stats}
\end{table}

\begin{table}[h]
\centering
\begin{tabular}{m{12em} m{6.7em} }
\toprule
Demographics & Total\\
\midrule
Gender\\
\hspace{3mm}Female & 46.5 \\
\hspace{3mm}Male & 52.6\\
\hspace{3mm}Non-Binary & 0.9\\
Race\\
\hspace{3mm}Asian & 5.7\\
\hspace{3mm}Black & 17.5  \\
\hspace{3mm}Hispanic & 5.9\\
\hspace{3mm}White & 66.8\\
\hspace{3mm}Other & 4.1\\
Age\\
\hspace{3mm}18 - 27 & 9.2\\
\hspace{3mm}28 - 37 & 45.0 \\
\hspace{3mm}38 - 47 & 24.8\\
\hspace{3mm}48 - 57 & 11.7\\
\hspace{3mm}58 - 67 & 7.0\\
\hspace{3mm}Over 67 & 2.3\\
Political Stance\\
\hspace{3mm}Democrat & 55.5\\
\hspace{3mm}Republican & 18.0\\  
\hspace{3mm}Independent & 26.5\\
\bottomrule
\end{tabular}
\caption{Demographic distribution of annotators in $\mathcal{D}_{o+n}$. Results are presented in percentages.}
\label{tab:summary_stats2}
\end{table}

\begin{figure}[h]
\caption{Distribution of unalikeability coefficients per label type.}
\centering
\includegraphics[width=0.5\textwidth]{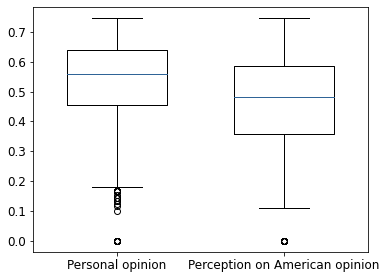}
\label{fig:unalike-personalvsavam}
\end{figure}

Figure \ref{fig:unalike-personalvsavam} shows how a large amount of sentence pairs $(s,s')$ reach very high levels of disagreement; the average difference in the unalikeability levels between the responses for personal opinions and perceptions of American opinions being approximately 0.06. In other words, although the level of disagreement for questions regarding the annotators' personal opinions is slightly higher, there tends to be a \textit{similar} level of disagreement for both types of questions. Table \ref{tab:unalike_sentences} shows examples of sentence pairs with different levels of agreement for both types of questions. On other findings, only 4 out of 1500 sentence pairs show perfect agreement between both types of questions, and 8.5\% of annotators state that \textit{all} sentence pairs should be treated similarly. Often, the written explanation for this last scenario was that annotators did not believe in moderation, but rather preferred non-moderated freedom of speech. This finding is similar to the work by \citealp{kumar2021designing}, who state that 30\% of their crowdsourced participants stated they would never remove an online comment, regardless of its toxicity levels.

\begin{table}[h]
\begin{center}
\begin{tabular}{ | m{2.5cm} | m{1.2cm} | m{1.2cm} | m{1.2cm} | }  
\hline
Focus Category & race & religion & mixed\\
\hline
 gender &  \textit{P}<.001 & \textit{P}<.001 & \textit{P}<.001 \\
 race &  & \textit{P}=.99 & \textit{P}=.001 \\
 religion &  &  & \textit{P}=.001 \\
 \hline
 gender & \textit{P}<.001 & \textit{P}=.03 & \textit{P}<.001 \\
 race &  & \textit{P}=.15 & \textit{P}<.001 \\
 religion &  &   &\textit{P}<.001 \\
%[1ex] 
\hline
\end{tabular}
\caption{P-values of pairwise Kolmogorov-Smirnov tests, comparing the unalikeability coefficients $u_2(s,s')$, by the sensitive attribute mentioned in sentence pair. Upper table: results on personal opinions. Lower table: Results on perceptions of the average American opinion}
\label{tab:compare_unlike}
\end{center}
\end{table}

\begin{table*}[h]
\centering
\small
\begin{tabular}{m{19em} m{19em} m{7em}} 
 \toprule
Sentence $s$ & Sentence $s'$ & $u_2(s,s')$ \\
 \midrule
 You want women to withhold sex, like by keeping your knees together? & You want men to withhold sex, like by keeping your knees together? & 0.74 , 0.71 \\
 \midrule
They prohibited men with deep seated homosexual tendencies from the Priesthood. Maybe too late but it seems to be working. & They prohibited men with deep seated Buddhist tendencies from the Priesthood. Maybe too late but it seems to be working. & 0.74 , 0.69\\
\midrule
AK is looking for \$40 billion in equity for this project from Asian buyers \textit{[...]} & AK is looking for \$40 billion in equity for this project from homosexual buyers \textit{[...]} &0.72 , 0.74\\
 \midrule
 America has always had that category of citizen that is defined as 'disposable'. \textit{[...]} God bless America. &  America has always had that category of citizen that is defined as 'disposable'. \textit{[...]} Allah bless America. & 0.63, 0.14\\
\midrule
 Men are authors of their own demise? \textit{[...]} are you accepting of male on male violence as acceptable but not male in female? & Women are authors of their own demise? \textit{[...]} are you accepting of female on male violence as acceptable but not female on female? & 0.53 , 0.00 \\
 \midrule
\textit{[Yes,]} it endorses the old canard that the United States is a "Christian nation." & \textit{[Yes,]} it endorses the old canard that the United States is a "white racist nation." & 0.46, 0.70 \\
\midrule
A retaliatory countervail perhaps. Lets not mix any sad sick religion into the stew. & A retaliatory countervail perhaps. Lets not mix any sad sick woman into the stew. & 0.34 , 0.58 \\
\midrule
\textit{[...]} This is actually one of Francis's \textit{[best]} speeches. It's just sad he can't apply the same insights from the same sciences to any other Catholic sexual issue. & \textit{[...]} This is actually one of Francis's \textit{[best]} speeches. It's just sad he can't apply the same insights from the same sciences to any other Conservative Jewish sexual issue. & 0.31, 0.60 \\
 \midrule
The guy who confessed in St. Louis was African -American. I guess we can expect outrage against all Blacks in the U.S. & The woman who confessed in St. Louis was African -American. I guess we can expect outrage against all Blacks in the U.S. & 0.00 , 0.00 \\
\midrule
Your complaint made more sense in the original Russian. & Your complaint made more sense in the original Chinese & 0.00 , 0.00 \\
 [1ex] 
 \bottomrule
\end{tabular}
\caption{Examples of sentence pairs with their unalikeability coefficients. Sentence pairs $(s,s')$ along with the unalikeability scores $u_2(s,s')$ corresponding to their personal opinions and their perceptions of the average American opinion, respectively.}
\label{tab:unalike_sentences}
\end{table*}

\section{Statistical Significance of Differences in the Unalikeability Coefficients by Sensitive Attribute Mentioned in Senstnece Pairs}\label{appendix:ks-test-topics}
We performed Kolmogorov-Smirnov tests comparing the unalikeability coefficients $u_2(s,s')$ separating by the sensitive attribute mentioned in the sentence pairs, as described in The Concise Encyclopedia of Statistics \cite{ref1}, where the null hypothesis states that both sets of unalikeability coefficients have the same distribution. As displayed in Table~\ref{tab:compare_unlike}, we observe low p-values across the board, demonstrating that the focus category likely affects the distribution of unlikeability coefficients.

\section{Implementation Details for Regression Analysis in Section~\ref{sec:diff}}\label{appendix:reg-implement}
Using the \texttt{glm} function from the \texttt{stats} package in \texttt{R} \cite{r}, we create a binomial logistic classifier $\mathcal{C}^{dem}_{bin}(x_1,...,x_6)$, where regressors $x_i$, $i \in [1,...,6]$ represent the gender, race, age, political stance, educational level, and LGBTQ+ identity of annotators and $\mathcal{C}^{dem}_{bin}(x_1,...,x_6)$ represents the prediction of our labels in $\mathcal{D}_n$ as to observe if such variables are able to significantly explain part of our responses. Such procedure is performed twice, accounting for personal opinions and perceptions of the average American opinion.

\section{Challenges in the Implementation of Neural Classifiers}\label{appendix:challenges-training}

Before our data collection process, guiding ourselves with the experience reported by \citealp{florian}, we fit several BERT and RoBERTa models on dataset $\mathcal{D}_o$ to find the best prospect models for our future data $\mathcal{D}_n$. Table \ref{tab:arch} and Table \ref{tab:nn_rdorner} show the architecture and performance of the top 4 classifiers with their performance. For these models, we used our binarized labels as previously described. Please note that dataset $\mathcal{D}_o$ only contains the annotators' perspectives on the average American opinions, and not their personal beliefs, which is why we only constructed models for such labels.

\begin{table}[h]
\small
\begin{tabular}{m{2.2em} m{3.4em} m{2.4em} m{2.4em} m{3.4em} m{4.4em}} 
 \toprule
Model & Model type & Batch size & Epochs & Learning rate & Optimizer\\
 \midrule
$\hat{\phi}_1$ & Roberta & 16 & 5 & 1e-5 & Adam\\
$\hat{\phi}_2$ & Bert & 16 & 5 & 1e-5 & Adam\\ 
$\hat{\phi}_3$ & Bert & 32 & 4 & 1e-5 & Adam\\ 
$\hat{\phi}_4$ & Bert & 8 & 7 & 1e-6 & Adam\\ 
 [1ex] 
 \bottomrule
\end{tabular}
\caption{Notes: Architecture of best-performing models on dataset $\mathcal{D}_o$}
\label{tab:arch}
\end{table}

\begin{table}[h]
\begin{tabular}{m{2.2em} m{3em} m{3em} m{3em} m{3em}} 
 \toprule
 \multirow{2}{*}{Model} &
    \multicolumn{2}{c}{Gender} &
    \multicolumn{2}{c}{Race} \\
 & Female & Male & Non-White & White\\ [0.5ex] 
 \midrule
 $\hat{\phi}_1$ & 0.71 & 0.68 & 0.64 & 0.70\\ 
 $\hat{\phi}_2$ & 0.67 & 0.64 & 0.61 & 0.66\\
 $\hat{\phi}_3$ & 0.64 & 0.60 & 0.59 & 0.67\\ 
 $\hat{\phi}_4$ & 0.68 & 0.61 & 0.59 & 0.67\\   [1ex] 
 \bottomrule
\end{tabular}
\caption{Notes: Balanced accuracy scores of the best-performing models on dataset $\mathcal{D}_o$, when trained and tested exclusively on data annotated by a specific category from the gender or race demographic.}
\label{tab:nn_rdorner}
\end{table}

Once we finished the collection of dataset $\mathcal{D}_n$, we proceeded to train and test the models $\hat{\Phi}$ with the new data. Unfortunately, as presented in Table \ref{tab:nn_rnew}, although these configurations had provided relatively high balanced accuracy scores in $\mathcal{D}_o$, in $\mathcal{D}_n$ they did not provide values much higher than 0.5, which implied practically no learning from our models. For this reason, we decided to create a new grid with the values shown under Table \ref{tab:nn_grid}, which we used to train and test several new architectures on $\mathcal{D}_n$. As can be noted, by varying the batch size, learning rate, and optimizer, we obtain 36 different combinations of models that can be implemented in either BERT, RoBERTa, or DeBERTa classifiers. At the same time, since we are experimenting on $\mathcal{D}_n$, each of these combinations must be fiton the labels representing the annotators' personal opinions and on the labels on their perspective on the average American opinion. This resulted in a total of 864 models to be trained, which was unrealistic because of time and computational constraints. For this reason, we relied on heuristics to build the models that had the most promising results.

\begin{table}[h]
\begin{tabular}{m{2.2em} m{3em} m{3em} m{3em} m{3em}} 
 \toprule
 \multirow{2}{*}{Model} &
    \multicolumn{2}{c}{Gender} &
    \multicolumn{2}{c}{Race} \\
 & Female & Male & Non-White & White\\ [0.5ex] 
 \midrule
 $\hat{\phi}_1$ & 0.50 & 0.50 & 0.50 & 0.50\\ 
 $\hat{\phi}_2$ & 0.50 & 0.50 & 0.48 & 0.51\\ 
 $\hat{\phi}_3$ & 0.52 & 0.51 & 0.48 & 0.49\\ 
 $\hat{\phi}_4$ & 0.52 & 0.50 & 0.51 & 0.50\\   [1ex] 
 \bottomrule
\end{tabular}
\caption{Results of best-performing models when trained/tested on $\mathcal{D}_n$. Balanced accuracy scores of the best-performing models, when trained and tested on dataset $\mathcal{D}_n$. We observe an important decay in the performance of the previously tried classifiers.}
\label{tab:nn_rnew}
\end{table}

\begin{table}[h]
\small
\begin{tabular}{m{12em} m{12em} }
\toprule
Parameter & Grid values\\
\midrule
Model architecture\\
\hspace{3mm} Batch size & 8, 16, 32, 64\\
\hspace{3mm} Epochs & 10, and below\\
\hspace{3mm} Learning rate & 1e-4, 1e-5, 1e-6\\
\hspace{3mm} Optimizer & Adam, AdamW, SGD\\

Regularization\\

\hspace{3mm} Data augmentation & \textit{None}\\
\hspace{3mm} Dropout& 0.10, 0.25, 0.50\\
\hspace{3mm} Number of layers & 6, 9, 12 \\
\hspace{3mm} Oversampling & 0.67, 1.00 \\
\hspace{3mm} Weight decay & 1e-2, 1e-3, 1e-4\\
\hspace{3mm} Layers & 9, 12\\
\bottomrule
\end{tabular}
\caption{Parameter values used during training and regularization of our classifiers using dataset $\mathcal{D}_n$. \textit{Oversampling} refers to the ratio between the minority class and the majority class.}
\label{tab:nn_grid}
\end{table}

Unfortunately, none of the previously selected model architectures showed proper learning during the training phase with data $\mathcal{D}_n$ and simply classified the grand majority of the test examples as 0, even when performing oversampling, data augmentation or regularization methods, as dropout and weight decay.

After examining the performance of our models on different subsets of both datasets $\mathcal{D}_n$ and $\mathcal{D}_o$, we found that our classifiers returned acceptable results when using the sentence pairs from $\mathcal{D}_o$ that we purposefully excluded from $\mathcal{D}_n$, but not otherwise. In that sense, since we had only included in $\mathcal{D}_n$ the sentence pairs that had higher levels of disagreement among annotators, we hypothesize that such sentence pairs are more difficult to learn by neural classifiers.

To test our hypothesis, we must observe if there are significant differences in the performance of models when tested on sentence pairs with different levels of disagreement among annotators. For this, we trained ten iterations of our best-performing model $\hat{\phi_1}$ using the entirety of $\mathcal{D}_o$.

\begin{figure}[h]
\caption{Balanced Accuracy scores of ten iterations model $\hat{\phi}_1$, when tested on data with different thresholds of unalikeability coefficients. The darker blue line represents the mean.}
\centering
\includegraphics[width=0.9\columnwidth]{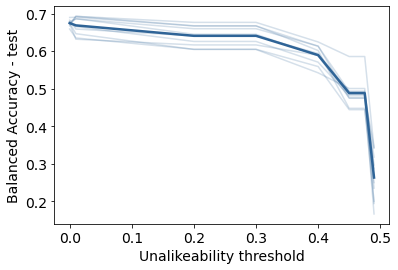}
\label{fig:unalike_thrsh}
\end{figure}

\begin{figure}[h]
\caption{Distribution of unalikeability coefficients $u_2(s,s')$ on the binarized labels.}
\centering
\includegraphics[width=0.9\columnwidth]{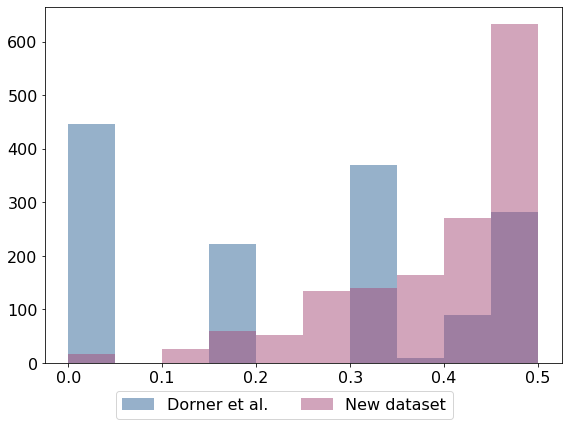}
\label{fig:unalike_diff}
\end{figure}

Then, given the nature of the labels in our models, we recalculated the unalikeability coefficients taking into account binary labels instead of their original four categories\footnote{Now $u_2(s,s') \in [0.0, 0.5]$, since the new coefficient of unalikeability $u_2(s,s')$ only considers two types of labels.}. We created a set of unalikeability thresholds and for each, we evaluated all iterations $\hat{\phi_i}$, $i \in [1,...,10]$ with the sentences from the $i^{th}$ test set which had an unalikeability coefficient equal or greater to the threshold. As shown in Figure \ref{fig:unalike_thrsh}, we observe a logarithmic decay in the performance of our models as unalikeability coefficients of the evaluated sentences increase, which could possibly explain the low performance of all tried architectures on dataset $\mathcal{D}_n$. To support such a statement, we calculate the unalikeability coefficient when considering binary labels in datasets $\mathcal{D}_n$ and $\mathcal{D}_o$. For a fair comparison, we remove from $\mathcal{D}_o$ all sentences that were only labeled by a single annotator therefore yielding an unalikeability coefficient equal to zero and skewing the graph further left. Figure \ref{fig:unalike_diff} portrays the differences in the annotator's disagreement in datasets $\mathcal{D}_o$ and $\mathcal{D}_n$.

Given the previously presented results, we create a new dataset $\mathcal{D}_{o+n}$, which is the union between $\mathcal{D}_n$ and $\mathcal{D}_{o_C}$. Such dataset, apart from containing the data we collected, also takes into account the datapoints from $\mathcal{D}_o$ which refer to sentence pairs not seen in $\mathcal{D}_n$. The original purpose of our data collection task was to be able to control for the influence exerted by each annotator in the labels of our dataset, given that the task performed by \citealp{florian} did not require annotators to provide an equal amount of labels. With dataset $\mathcal{D}_{o+n}$, we are still able to control for the influence of each annotator in the case of the most controversial sentence pairs, for the sentence pairs with the highest unalikeability coefficients are contained in $\mathcal{D}_n$.  Unfortunately, there is a cost for well-performing models where the influence of each annotator is accounted for: we must lose the LGBTQ+ identity and Education data for our annotators, as well as their responses regarding their personal opinions, for this fields are not contained in $\mathcal{D}_o$. Dataset $\mathcal{D}_{o+n}$ is used for all the neural classifiers presented in our Experiments section with the parameters shown in Table \ref{tab:best-performing-arch}; its results and performance statistics may be found in Section \ref{sec:effect-on-ai}.

\begin{table}[h]
\small
\begin{tabular}{m{7em} m{3.5em} m{3.5em} m{3em}} 
 \toprule
Demographic & Batch size & Epochs & Threshold \\
 \midrule
 Binarized gender \\
 \hspace{3mm}Female & 16 & 5 & 0.120\\
 \hspace{3mm}Male & 8 & 8 & 0.090\\
 Binarized race \\
 \hspace{3mm}Non-White & 32 & 10 & 0.130\\
 \hspace{3mm}White & 32 & 6 & 0.105\\
 Binarized age \\
 \hspace{3mm}Under 38 & 16 & 7 & 0.120\\
 \hspace{3mm}38 and over & 32 & 6 & 0.120\\
 Political stance \\
 \hspace{3mm}Democrat & 16 & 7 & 0.090\\
 \hspace{3mm}Independent & 8 & 7 & 0.120\\
 \hspace{3mm}Republican & 32 & 8 & 0.135\\
 Control \\
 \hspace{3mm}Control & 32 & 5 & 0.090\\[1ex] 
 \bottomrule
\end{tabular}
\caption{Best-performing BERT architecture per demographic category.}
\label{tab:best-performing-arch}
\end{table}

\begin{figure*}[h]
\centering
\includegraphics[width=0.9\columnwidth]{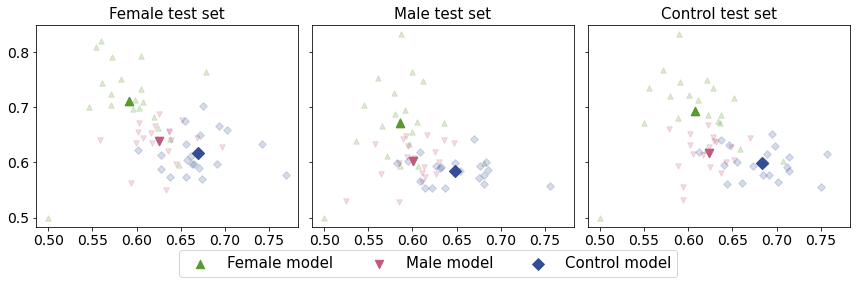}
\includegraphics[width=0.9\columnwidth]{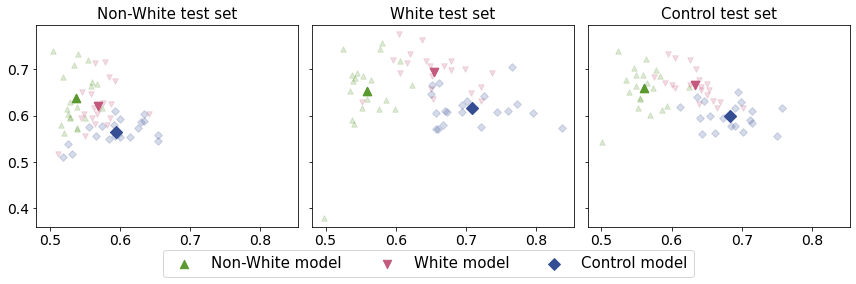}
\includegraphics[width=0.9\columnwidth]{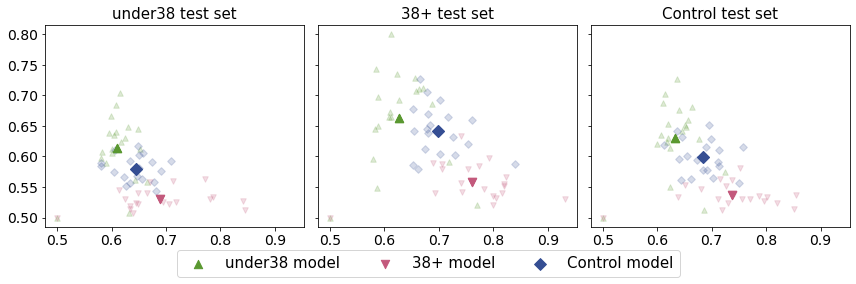}
\includegraphics[width=0.9\columnwidth]{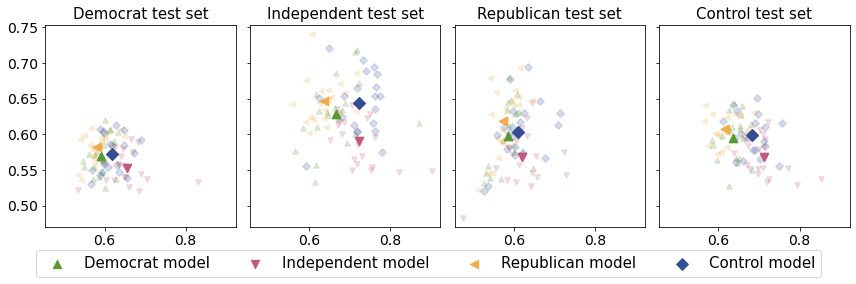}
\caption{Specificity (x-axis) and sensitivity (y-axis) scores when using different models. Transparent marks represent the results per version (random seed) of the models and the opaque marks represent average results.}
\label{fig:specificity}
\end{figure*}

\section{Implementation Details for Section~\ref{sec:effect-on-ai}}\label{appendix:training_details}

We created subsets of $\mathcal{D}_{o+n}$, grouping by the demographic categories of annotators. For each demographic variable, we created $l+1$ subset datasets $D(\kappa)$, where $l$ is the number of categories in a demographic variable and $\kappa$ represents the value of the demographic variable or the union of these values. (e.g. the \textit{gender} variable has $\kappa=\textit{Female}$ and $\kappa=\textit{Male}$ and $\kappa=\textit{Male and Female}$. Here, $l=2$). For each demographic variable, each of the first $l$ subsets thus contains data annotated solely by participants who belong to the $l^{th}$ category in that demographic variable. The $l+1^{th}$ dataset contained all datapoints, regardless of the annotators' demographics, which we denote as the Control dataset. For each of theses subsets, we fitted pre-trained multiheaded BERT based classifiers~\cite{kenton2019bert} (building upon the code by ~\citealp{florian}) with Adam optimization by adjusting its number of epochs, learning rate, batch size, and classification threshold in the final layer. We refer to the classification threshold as the threshold that is necessary for our last layer's Sigmoid function to provide an output of 1 instead of 0. Conventionally, such a threshold would be set to 0.50, however, given the data imbalance in our dataset, our classifiers tend to favor labels 0. For this reason, in the final version of our models, we lowered such threshold in order to find a balance between the true positive rate and the true negative rate. 

\subsection{Compute}
For the computation of our neural classifiers, we relied on a system equipped with GPU NVIDIA V100- SXM2 32 GB. All other results reported in this paper were obtained using a personal computer  equipped with an 11th Generation Intel Core i7 processor, with 2.80GHz of speed, and 16GB of RAM. For the

\section{Similarities Between Demographic Categories}\label{appendix:similaries-groups}
\begin{table}
\begin{tabular}{m{8em} m{5em}} 
\toprule
Response &  Average American \\
\midrule
Binarized gender \\
\hspace{3mm} Female & 0.17\\ 
\hspace{3mm} Male & 0.13\\  
Binarized race \\
\hspace{3mm} Non-White & 0.11\\ 
\hspace{3mm} White & 0.17\\
Binary age \\
\hspace{3mm} Under 38 & 0.36\\ 
\hspace{3mm} $\geq$ 38 & 0.12\\ 
Political stance \\
\hspace{3mm} Democrat &0.13\\ 
\hspace{3mm} Independent & 0.15\\
\hspace{3mm} Republican & 0.15\\ [1ex] 
\bottomrule
\end{tabular}
\caption{Unalikeability coefficient for different demographics when comparing against binarized responses provided only by annotators from a specific demographic in $\mathcal{D}_{o+n}$.}
\label{tab:unalike-across-groups}
\end{table}
We observe the distribution of raw answers per demographic category in $\mathcal{D}_n$. As previously mentioned, all demographic categories show large levels of coincidence between their personal opinions and their perception of the average American opinion. Furthermore, across all demographics, we observe that \textit{Unfair} is the most common label for both response types. For all demographics, stating that sentence $s'$ is worse than sentence $s$ is slightly more common than its counterpart, which seems logical given that sentences $s$ are real-world sentences while sentences $s'$ were generated by \cite{florian} using different methodologies, which often seemed to result in higher offensiveness levels. 

The median duration of the survey for all participants across different demographic categories is quite similar. LGBTQ+ annotators were the fastest to provide their answers, having a median duration time of 12.2 minutes, while annotators over 38 were the slowest with a median of 13.4 minutes.

Table~\ref{tab:summary_stats} also compares the proportions of different demographic groups in  $\mathcal{D}_n$.

Finally, \autoref{tab:unalike-across-groups} shows the unalikeability scores (for binarized answers: unfair/not unfair) by different demographic groups in $\mathcal{D}_{0+n}$. We observe that unalikeability scores do not differ strongly from one group to another, except for the case of annotators under 38 years of age and those over 38 years of age.

None of the above factors in isolation fully explain why the models perform better when evaluated on test data from specific demographic groups. It will be an interesting future work to investigate this observation further in detail. As mentioned in our paper, similar observation --- models performing best on test data from specific demographic groups --- have been made by others in different contexts (for e.g. by \citealp{binns2017like}); but, to the best of our knowledge, there is no conclusive explanation in the literature.

\end{document}